\documentclass{article} % For LaTeX2e
\usepackage{iclr2026_conference,times}

\usepackage{amsmath,amsfonts,bm}

\def\eqref#1{equation~\ref{#1}}
\def\1{\bm{1}}

\DeclareMathAlphabet{\mathsfit}{\encodingdefault}{\sfdefault}{m}{sl}
\SetMathAlphabet{\mathsfit}{bold}{\encodingdefault}{\sfdefault}{bx}{n}

\usepackage{hyperref}
\usepackage{url}
\usepackage{graphicx}

\usepackage{algorithm}
\usepackage{algorithmic}

\usepackage{booktabs, multirow} % For formal tables
\usepackage{xspace}
\usepackage{color, xcolor,colortbl} 
\usepackage{amsmath}
\usepackage{tikz}
\usepackage{subfigure}
\usepackage{caption}
\usepackage{wrapfig}
\usepackage{tabularx}
\usepackage{graphicx}

\usepackage{multirow}
\usepackage{colortbl}
\usepackage{subcaption}
\definecolor{skyblue}{RGB}{135,206,235} 
\definecolor{lightgreen}{RGB}{144,238,144} % 浅绿色
\definecolor{darkgreen}{RGB}{0,100,0}     % 深绿色
\usepackage[dvipsnames]{xcolor}
\usepackage{pifont}

\usepackage[most]{tcolorbox}

\newtcolorbox{promptbox}[1]{
    colback=gray!5!white, % 背景颜色
    colframe=gray!35!white, % 边框颜色
    fonttitle=\bfseries,
    coltitle=black,
    title=#1,
    sharp corners,
    boxsep=4pt,
    left=4pt,
    right=4pt,
    top=2pt,
    bottom=4pt
}

\title{Beyond Isolated Facts: Synthesizing Narrative and Grounded Supervision for VideoQA}
\author{Jianxin Liang$^{1}$, Tan Yue$^{1}$, Yuxuan Wang$^{1}$, Yueqian Wang$^{1}$,\\
\textbf{Zhihan Yin}$^{1}$, \textbf{Huishuai Zhang}$^{1,*}$, \textbf{Dongyan Zhao}$^{1,2,}$\thanks{Corresponding Authors Huishuai Zhang and Dongyan Zhao} \\
$^{1}$ Wangxuan Institute of Computer Technology, Peking University \\
$^{2}$State Key Laboratory of General Artificial Intelligence \\
\texttt{\{liangjx,zhanghuishuai,zhaody\}@pku.edu.cn}
}

\iclrfinalcopy % Uncomment for camera-ready version, but NOT for submission.
\begin{document}

\maketitle

\begin{abstract}
    The performance of Video Question Answering (VideoQA) models is fundamentally constrained by the nature of their supervision, which typically consists of isolated, factual question-answer pairs. This "bag-of-facts" approach fails to capture the underlying narrative and causal structure of events, limiting models to a shallow understanding of video content. To move beyond this paradigm, we introduce a framework to synthesize richer supervisory signals. We propose two complementary strategies: Question-Based Paraphrasing (QBP), which synthesizes the diverse inquiries (what, how, why) from a video's existing set of question-answer pairs into a holistic narrative paragraph that reconstructs the video's event structure; and Question-Based Captioning (QBC), which generates fine-grained visual rationales, grounding the answer to each question in specific, relevant evidence. Leveraging powerful generative models, we use this synthetic data to train VideoQA models under a unified next-token prediction objective. 
    Extensive experiments on STAR and NExT-QA validate our approach, demonstrating significant accuracy gains and establishing new state-of-the-art results, such as improving a 3B model to 72.5\% on STAR (+4.9\%) and a 7B model to 80.8\% on NExT-QA. Beyond accuracy, our analysis reveals that both QBP and QBC substantially enhance cross-dataset generalization, with QBP additionally accelerating model convergence by over 2.5x.    These results demonstrate that shifting data synthesis from isolated facts to narrative coherence and grounded rationales yields a more accurate, efficient, and generalizable training paradigm.
    % Extensive experiments on STAR and NExT-QA validate our approach. Our method improves a 3B model's accuracy on STAR to 72.5\% (+4.9\%) and pushes a 7B model to a new state-of-the-art of 80.8\% on NExT-QA. These results demonstrate that shifting the focus of data synthesis from isolated facts to narrative coherence and grounded rationales not only boosts state-of-the-art accuracy but also significantly improves model efficiency and generalization.
\end{abstract}

\section{Introduction}

Video Question Answering (VideoQA)~\citep{patel2021recent,zhong2022Video} is a pivotal multimodal task that requires models to reason over complex visual and textual inputs to answer natural language questions about videos. 
%Its importance spans a wide range of applications, from video retrieval and surveillance~\citep{sreenu2019intelligent} to assistive technologies and interactive AI systems~\citep{rajavel2022iot}. 
It has broad applications, including video retrieval and surveillance~\citep{sreenu2019intelligent}, as well as assistive technologies and interactive AI systems~\citep{rajavel2022iot}.
% However, progress in VideoQA remains constrained by the scarcity and limitations of existing annotated datasets. 

Benefiting from recent advances in Large Language Models (LLMs)~\citep{achiam2023gpt,openai2024gpt4o} and Multimodal Large Language Models (MLLMs)~\citep{team2023gemini,openai2024gpt4v}, VideoQA has made rapid progress. Strong MLLMs equipped with cross-modal attention, temporal modeling, and instruction-following abilities have substantially improved accuracy on standard benchmarks. 
% Despite continuous efforts to advance VideoQA through stronger architectures and auxiliary tasks~\citep{liang2024end,liang-etal-2025-reasvqa,zhang2023video}, the effectiveness of current models is fundamentally tied to the quality of supervision they receive. 
% Nevertheless, significant challenges remain. Conventional datasets~\citep{jang2017tgif,xiao2021next,wu2021star} are often trimmed around scene boundaries, yielding simplified plots that omit subtle movements or fine-grained interactions. This lack of detail frequently causes models to hallucinate when required to provide precise or causal explanations~\citep{zhang2024video}. Moreover, training on such datasets encourages models to overfit to narrow domain patterns, which not only limits generalization but can even degrade performance when transferred to other benchmarks.
Nevertheless, significant challenges remain, rooted in the very structure of our training data. Conventional datasets~\citep{jang2017tgif,wu2021star,xiao2021next} are composed of discrete question-answer pairs that, while factually correct, present video content as a series of fragmented, isolated facts. This format omits the rich web of inter-dependencies, such as the causal, temporal, and social links, that connect these facts into a coherent event. 
To highlight this fundamental limitation, Table~\ref{tab:intro_example} presents a typical set of human-annotated questions for a single video.
%To illustrate this fundamental issue, consider the typical set of human-annotated questions for a single video shown in Table~\ref{tab:intro_example}.

Individually, each QA pair in Table~\ref{tab:intro_example} provides a useful, atomic piece of information. However, their true value lies in the semantic links that are entirely ignored by conventional training paradigms. For instance, understanding why the people are resting (Q3, Q5) is contingent on knowing they are on snowmobiles (Q1). Inferring their relationship as 'friends' (Q6) is not a direct visual fact but an inference supported by the playful 'posing' interaction (Q4). Current models~\citep{ko2023large, liang2024end}, trained on this data, are tasked with learning from a "bag-of-facts," forcing them to rely on shallow correlations rather than deep, structural understanding. This not only limits generalization but is a primary cause of model hallucination when complex reasoning is required. The critical research gap, therefore, is not just the scarcity of data, but the absence of a supervision signal that represents this underlying event structure.
% Previous attempts at addressing data scarcity have operated at a surface level. Methods like generating new QA pairs or applying unimodal augmentations primarily increase the volume of isolated, factual annotations~\cite{grunde2021agqa}. While beneficial, this paradigm reinforces a fundamental limitation: it trains models to retrieve discrete facts rather than to comprehend structured events. Each QA pair remains a fragmented piece of supervision, failing to provide the connective tissue—the causal and temporal logic—that links individual facts into a coherent understanding. What is missing is not merely more data, but a new form of supervision that can represent the underlying event structure itself.
% Previous attempts at addressing data scarcity include generating additional QA pairs~\cite{grunde2021agqa,} or applying unimodal augmentations such as visual transformations or paraphrasing at the single-question level~\cite{grunde2021agqa}. While these methods bring incremental benefits, they fail to capture the semantic richness of the video content and typically treat each QA pair as an isolated supervision signal. What is missing is a form of supervision that goes beyond fragmented annotations to provide models with a more coherent and contextually grounded understanding of the underlying events.

To address this fundamental challenge,
% To overcome these challenges, 
we propose a framework that introduces two novel forms of supervision by transforming the fragmented QA pairs already present in existing datasets. 
% To overcome these challenges, we propose a framework that introduces two novel forms of supervision. 
Our first strategy, \textbf{Question-based Paraphrasing (QBP)}, addresses the need for structured understanding. It leverages the rich interrogative diversity (what, how, why) inherent in human-annotated questions to reverse-engineer a video's underlying event structure. Instead of treating them as a bag of isolated facts, QBP compels a LLM to synthesize these descriptive, procedural, and causal inquiries into a single, logic-infused narrative. This process transforms fragmented seeds of human curiosity into a holistic, narrative-level supervision signal. However, a global narrative alone cannot guarantee visual grounding. To this end, our second strategy, \textbf{Question-based Captioning (QBC)}, provides instance-level grounding. It generates fine-grained, question-conditioned captions that serve as visual rationales, forcing the model to anchor its reasoning in specific, relevant visual evidence. Together, QBP and QBC provide two orthogonal yet synergistic forms of supervision: one that builds a coherent narrative fabric, and another that ties each thread of that fabric to a concrete visual detail.
% To overcome these challenges, we propose two complementary data synthesis strategies that enrich VideoQA supervision with diverse, high-quality textual signals. The first, Question-based Paraphrasing (QBP), aggregates multiple QA pairs associated with the same video into holistic narrative paragraphs. By capturing inter-question dependencies and temporal or causal structure, it provides a global, narrative-level supervision signal that is otherwise missing in raw QA datasets. However, global narratives alone cannot ensure that models remain well-grounded in the visual content. To this end, we introduce QBC, which generates fine-grained, question-conditioned captions that highlight visual details most relevant to answering each query. Whereas QBP enforces global coherence across multiple questions, QBC enforces local grounding between individual queries and their supporting evidence. Together, they provide two orthogonal yet synergistic forms of synthetic supervision: narrative-level coherence and instance-level grounding.

Extensive experiments validate the effectiveness of our approach. On two widely used benchmarks, NExT-QA and STAR~\citep{xiao2021next,wu2021star}, our QBP+QBC strategies consistently improve performance across different model backbones. For example, with a Qwen2.5-VL-3B~\citep{bai2025qwen25vltechnicalreport} backbone, accuracy on STAR improves from 67.6\% to 72.5\%, a gain of nearly +5 points. Larger backbones like Qwen2.5-VL-7B and MiMo-VL-SFT~\citep{coreteam2025mimovltechnicalreport} also benefit, with our QBP+QBC supervision pushing a 7B model to a new state-of-the-art of 80.8\% on NExT-QA. Beyond raw accuracy, our analyses reveal significant secondary benefits: QBP's narrative supervision accelerates model convergence by more than 2.5 times, while both strategies substantially improve cross-dataset generalization, demonstrating enhanced robustness.

In summary, our contributions are as follows:
(i) We propose a new supervision paradigm for VideoQA that moves beyond isolated facts, introducing two complementary synthesis strategies (QBP and QBC) to generate narrative-level and instance-level supervision.
(ii) We demonstrate through large-scale experiments that our framework significantly improves both in-domain accuracy and cross-dataset generalization, achieving new state-of-the-art results on multiple challenging benchmarks.
(iii) We provide a comprehensive analysis of the distinct benefits of our methods, showing that QBP accelerates model convergence by over 2.5x while both strategies enhance generalization, underscoring the efficiency and robustness of our approach.
% (i) we propose two complementary data synthesis strategies, QBP and QBC, that transform fragmented annotations into richer supervision;
% (ii) we demonstrate through large-scale experiments that these methods significantly improve both in-domain accuracy and cross-dataset generalization, achieving new state-of-the-art results; and
% (iii) we provide detailed analyses showing that QBP accelerates convergence while QBC scales gracefully, underscoring their efficiency and robustness.

\section{Related Work}

\paragraph{Video Question Answering: From Architectures to Data Bottlenecks.}
VideoQA is a challenging multimodal task requiring complex spatio-temporal reasoning. Early progress was largely driven by architectural innovations, from spatio-temporal attention mechanisms~\citep{xu2017video,jang2017tgif} and graph-based models~\citep{xiao2022video} to large-scale pre-trained transformers~\citep{yang2020bert,wang2022internvideo}. While these models have become increasingly sophisticated, their performance is fundamentally bottlenecked by the available training data~\citep{zhang2023video, li2023videochat}. Manually annotating large-scale, diverse, and unbiased datasets that cover complex reasoning scenarios is prohibitively expensive. Consequently, the field's focus is gradually shifting from purely architectural improvements to data-centric approaches~\citep{liang-etal-2025-reasvqa} that can enhance the quality and form of the supervision signal itself.

\paragraph{Data Synthesis for Video Understanding.}
Early approaches in VideoQA relied on rule-based templates~\citep{grunde2021agqa,wu2021star} or simple question generation~\citep{falcon2020data}, but these methods often produce syntactically simple and semantically repetitive data. The advent of powerful generative models has enabled more sophisticated synthesis. MLLMs like Video-LLaMA~\citep{zhang2023video} can generate descriptive video captions, while recent work such as LLaVA-Video~\citep{zhang2024video} and ShareGPT4V~\citep{chen2024sharegpt4v} has prompted LLMs like GPT-4~\citep{achiam2023gpt} to generate a variety of video-centric textual data. 

\begin{table}[t]
\centering
\caption{An example of fragmented yet semantically linked QA pairs for a single video from NExT-QA. While each pair provides an isolated fact, their inter-dependencies (rightmost column) reveal a richer event structure. Conventional training paradigms ignore these crucial links, forcing models to learn from a "bag-of-facts" and hindering deep reasoning.}
\label{tab:intro_example}
% 使用 tabularx 环境，并指定总宽度为 \textwidth
% 列定义：第一列和第三列使用可自动换行和扩展的 X 类型，第二列使用不换行的 l 类型
% >{\raggedright\arraybackslash} 是为了让换行的文本左对齐，而不是两端对齐，这样在窄列中更美观
\resizebox{0.99\linewidth}{!}{
\begin{tabularx}{1.3\textwidth}{ >{\raggedright\arraybackslash}X | l | >{\raggedright\arraybackslash}X }
\toprule
\textbf{Question \& Answer} & \textbf{Question Type} & \textbf{Semantic Links \& Implied Context} \\
\midrule
Q1: How are the people transported on snow? (snowmobile) & Transportation & Context for understanding the setting and actions in Q3, Q5. \\
\hline
Q2: What is the weather like? (cold) & Scene / Weather & Provides general atmospheric context for the entire scene. \\
\hline
Q3: Why is the person in red sitting on a snowmobile? (resting) & Action Reasoning & Links the action (`sitting`) to a purpose (`resting`), dependent on Q1, Q5. \\
\hline
Q4: How does the man in black react to the camera? (poses) & Interaction & Implies a social relationship (`friends`, Q6) and connects to camera actions (Q7, Q8). \\
\hline
Q5: Why are the snowmobiles parked? (resting) & Causal Reasoning & The overarching reason for the scene's static nature, connects to Q1, Q3. \\
\hline
Q6: What is the relationship between the people? (friends) & Social Relation & Inferred from playful interactions like `posing` (Q4) and `taking photos` (Q7, Q8). \\
\hline
Q7: Why is the man in blue holding a camera? (to take a photo) & Action Purpose & Explains the core interaction, directly linked to the reaction in Q4 and action in Q8. \\
\hline
Q8: What does the man in red do? (takes a photo) & Specific Action & A key interaction that supports the inference of `friends` (Q6) and explains the `posing` (Q4). \\
\bottomrule
\end{tabularx}}
\end{table}

However, despite the improved quality, the dominant paradigm remains the generation of more isolated data points—be it captions or individual QA pairs. This approach enriches the dataset in volume but fails to address the core problem we identify in our introduction: the structural fragmentation of supervision. These methods do not provide the connective tissue that links discrete facts into a coherent event structure, which is essential for deep reasoning.

\paragraph{Our Contribution in Context.}
Our work is situated within this trend of LLM-based data synthesis but makes a distinct and complementary contribution. Instead of generating \textit{more} fragmented data, we focus on creating \textit{new forms} of structured supervision. We are the first to propose a dual-pronged framework that explicitly addresses the structural deficit. Our Question-based Paraphrasing introduces a novel narrative-level supervision signal, designed to reconstruct the video's event structure from existing queries. Concurrently, our Question-based Captioning provides rationale-level supervision, forcing a tight, evidence-based alignment between a specific query and its visual proof. By synthesizing these two synergistic forms of supervision, our work directly tackles the limitations of the "bag-of-facts" paradigm that characterizes prior work.

\section{Method}

\begin{figure}[t]
    \centering
    \includegraphics[width=\linewidth]{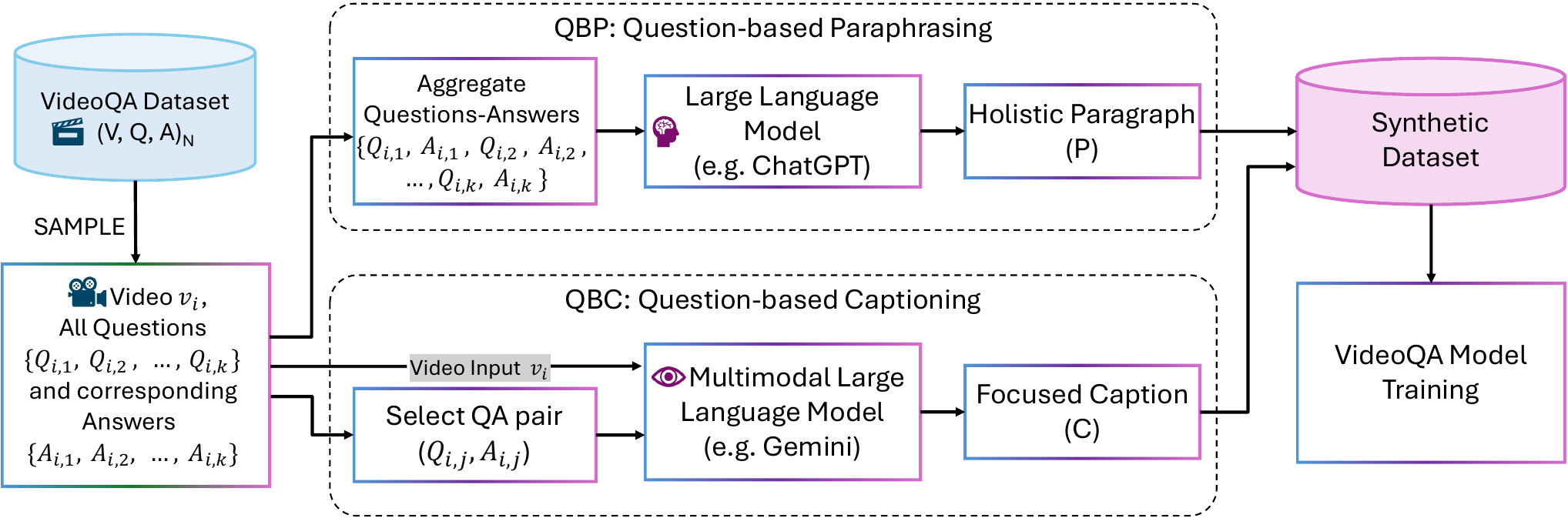}
    % \caption{An overview of our synthetic data generation pipeline. The process starts with an original VideoQA dataset and branches into two independent methods: Question-based Paraphrasing (QBP), which generates a holistic paragraph from multiple questions, and Question-based Captioning (QBC), which creates a focused caption from a single question and the video. The outputs are combined to form a synthetic dataset for downstream model training.}
    \caption{An overview of our framework for transforming fragmented QA pairs into structured supervision. Question-based Paraphrasing (QBP) synthesizes multiple QA pairs into a holistic narrative for global context, while Question-based Captioning (QBC) generates a visual rationale from a single QA pair to provide fine-grained, evidence-based grounding.}
    \label{fig:qbp}
\end{figure}

Our work introduces a novel framework for synthesizing high-quality training data to improve VideoQA models. Instead of simply augmenting existing datasets, we propose a method to transform the sparse, fragmented supervision inherent in human-annotated QA pairs into dense, multi-level training signals. We develop two complementary synthesis techniques: Question-based Paraphrasing (QBP), which generates holistic, narrative-level supervision; and Question-based Captioning (QBC), which provides fine-grained, instance-level grounding. The overall pipeline of our method is illustrated in Figure~\ref{fig:qbp}.

% Our goal is to improve VideoQA models by leveraging automatically synthesized training data. Our framework repurposes the rich but fragmented signals within existing datasets—specifically, the diverse, human-annotated questions associated with each video. We propose two complementary strategies: Question-based Paraphrasing (QBP), which generates holistic, narrative-level supervision; and Question-based Captioning (QBC), which provides fine-grained, instance-level grounding.
% We propose two complementary data synthesis strategies: \textbf{Question-based Paraphrasing (QBP)} and \textbf{Question-based Captioning (QBC)}. Both approaches exploit the inherent structure of VideoQA datasets, where each video is accompanied by a diverse set of questions, to construct additional supervision signals that extend beyond the original annotations. 

\subsection{Problem Formulation}

Formally, let the source data be a collection of videos $\mathcal{V} = \{v_i\}_{i=1}^N$. Each video $v_i$ is associated with a set of $K_i$ human-annotated question-answer pairs, which we denote as a question group $\mathcal{G}_i = \{(Q_{i,k}, A_{i,k})\}_{k=1}^{K_i}$. A video $v_i$ is represented as a sequence of $T$ uniformly sampled frames:
\[
v_i = \{ f_{i,1}, f_{i,2}, \dots, f_{i,T} \}, \quad f_{i,t} \in \mathbb{R}^{H \times W \times 3}.
\]
Our first step is to leverage the question groups $\{\mathcal{G}_i\}_{i=1}^N$ to synthesize two new datasets derived from the videos in $\mathcal{V}$:
\begin{itemize}
    \item A narrative-level dataset, $\mathcal{D}^{\text{QBP}} = \{(v_i, \tilde{d}_i^{\text{narrative}})\}_{i=1}^N$, generated via our QBP strategy.
    \item A rationale-level dataset, $\mathcal{D}^{\text{QBC}} = \bigcup_{i,k} \{(v_i, \tilde{d}_{i,k}^{\text{rationale}})\}$, generated via our QBC strategy.
\end{itemize}
Crucially, our training paradigm for a model $\mathcal{M}$ does not use the original, fragmented QA pairs for supervision. Instead, our objective is to train $\mathcal{M}$ exclusively on the union of our synthesized datasets:
\[\mathcal{D}_{\text{train}} = \mathcal{D}^{\text{QBP}} \cup \mathcal{D}^{\text{QBC}}.\]
We aim to demonstrate that training on $\mathcal{D}_{\text{train}}$ yields superior performance in terms of reasoning, generalization, and grounding compared to models trained on the standard QA dataset format.

\subsection{Question-based Paraphrasing (QBP): Building Global Narratives}

A key limitation of the standard VideoQA training paradigm is its reliance on isolated question-answer pairs as supervision units. This fragmentation ignores the rich semantic dependencies that often exist between questions associated with the same video. As illustrated in Table~\ref{tab:intro_example}, questions may be temporally, causally, or logically linked (e.g., Q1, Q3, and Q5 all concern the resting state, while Q7 and Q8 both hinge on camera-related actions). Ignoring these relations prevents models from forming a unified representation of the video's event structure.

\textbf{Conceptual Framework.} To overcome this fragmentation, we introduce Question-based Paraphrasing (QBP), a strategy designed to reconstruct the underlying event structure from these isolated annotations. Our key insight is that the set of human-annotated questions for a video, $\mathcal{G}_i$, is not a random collection of facts, but a rich sample of \textit{interrogative diversity}. These questions probe the video's content at multiple semantic levels: 'what' questions establish static entities, 'how' questions trace dynamic processes, and 'why' questions uncover causal relationships.

QBP frames the data synthesis task as a \textit{reasoning integration} problem. It compels an LLM to move beyond answering individual questions and instead synthesize these descriptive, procedural, and causal inquiries into a single, logic-infused narrative. This process transforms the fragmented ``bag-of-facts'' represented by $\mathcal{G}_i$ into a holistic, narrative-level supervision signal, $\tilde{d}_i^{\text{narrative}}$. By training on these narratives, the model is exposed to the connective tissue of the event, encouraging a shift from simple fact retrieval to structured event comprehension.

\textbf{Formalization.} Formally, given the question group $\mathcal{G}_i$ for a video $v_i$, we employ a LLM, denoted as $\Phi_{\text{QBP}}$, to generate a single narrative description $\tilde{d}_i^{\text{narrative}}$:
\[
\tilde{d}_i^{\text{narrative}} = \Phi_{\text{QBP}}(\mathcal{G}_i).
\]
% --- Start of ADDED SENTENCE ---
Here, the full question-answer pairs in $\mathcal{G}_i$ are provided as input, allowing the LLM to use the ground-truth answers as factual cornerstones for its narrative reconstruction.
% --- End of ADDED SENTENCE ---
The prompt for $\Phi_{\text{QBP}}$ explicitly instructs the model to integrate information across all QA pairs in $\mathcal{G}_i$ into a coherent, fluent paragraph, capturing latent dependencies. This process is applied to all videos in the source collection to construct the full narrative-level dataset:
\[
\mathcal{D}^{\text{QBP}} = \{(v_i, \tilde{d}_i^{\text{narrative}})\}_{i=1}^N.
\]

\subsection{Question-based Captioning (QBC): Enhancing Visual Grounding}

While QBP provides models with a global narrative context, a persistent challenge in VideoQA is \textit{visual grounding}: ensuring answers are derived from tangible visual evidence rather than dataset biases or spurious correlations. Models often fail at fine-grained spatio-temporal localization, particularly for complex ``why'' or ``how'' questions. For example, given the question ``Why did the person drop the ball?'', a generic caption like ``A person is playing with a ball'' offers little explanatory power. In contrast, a targeted \textit{visual rationale} such as ``The person’s hand slips as they try to catch the ball, causing it to fall'' directly links the reasoning to an observable, causal event.

\textbf{Conceptual Framework.} To instill this level of grounding, we propose Question-based Captioning. This strategy generates fine-grained visual rationales conditioned on individual question-answer pairs. The question focuses the general topic, while the ground-truth answer provides a specific anchor for correctness. This prompts a Multimodal Large Language Model to identify and describe the precise visual evidence that \textit{justifies} the given answer. This process creates a strong alignment between a query, its correct answer, and its visual proof, forcing the downstream model to learn not just \textit{what} the answer is, but \textit{why} it is correct based on the video.

\textbf{Formalization.} For each video $v_i$ and each of its associated question-answer pairs $(Q_{i,k}, A_{i,k})$ from the question group $\mathcal{G}_i$, we synthesize a targeted visual rationale $\tilde{d}_{i,k}^{\text{rationale}}$. This is generated by a Multimodal LLM, denoted as $\Phi_{\text{QBC}}$, which takes the video, the question, and the answer as input:
\[
\tilde{d}_{i,k}^{\text{rationale}} = \Phi_{\text{QBC}}(v_i, Q_{i,k}, A_{i,k}).
\]
Here, explicitly providing the ground-truth answer $A_{i,k}$ is a crucial design choice. It constrains the generation task, ensuring the correctness and relevance of the output. Instead of open-endedly describing the scene, the MLLM is instructed to find and articulate the specific visual evidence that supports the given correct answer. The prompt is carefully designed to forbid the model from merely repeating the answer, forcing it to generate a descriptive proof. This synthesis is performed for all question-answer pairs in the original dataset to construct the rationale-level dataset:
\[
\mathcal{D}^{\text{QBC}} = \bigcup_{i=1}^N \bigcup_{k=1}^{K_i} \{(v_i, \tilde{d}_{i,k}^{\text{rationale}})\}.
\]
This dataset consists of `(video, text)` pairs, structurally identical to $\mathcal{D}^{\text{QBP}}$, where each text serves as a grounded explanation for an implicit question-answer pair.

\textbf{Complementary Nature.} Conceptually, QBC complements QBP. Whereas QBP focuses on constructing holistic narratives that capture global dependencies across multiple questions, QBC operates at a fine-grained level, enforcing a tight alignment between an individual query-answer pair and its supporting visual evidence. Together, they provide two orthogonal yet synergistic forms of synthetic supervision, namely global narrative coherence and local visual grounding, which constitute our final training set $\mathcal{D}_{\text{train}} = \mathcal{D}^{\text{QBP}} \cup \mathcal{D}^{\text{QBC}}$.

\section{Experiments}

In this section, we conduct comprehensive experiments to empirically validate our proposed data synthesis framework. Our evaluation is structured to answer several key questions regarding its effectiveness, properties, and the quality of its outputs. 
% We begin by detailing our experimental setup, including an analysis of the seed datasets and the statistical properties of our synthetic data. We then present the main performance evaluation against state-of-the-art models. This is followed by in-depth component-wise analyses of the distinct properties of QBP and QBC, covering aspects like cross-dataset generalization, the impact of seed data diversity, convergence speed, and data scaling. Finally, we provide a rigorous human evaluation to assess the quality and factual fidelity of our generated supervision signals.
We first describe our experimental setup and then present a human evaluation to assess the quality and factual fidelity of the generated supervision signals, including an analysis of the seed datasets and the statistical properties of our synthetic data. Finally, we evaluate model performance and analyze the contributions of different components in our framework.

\textbf{Training.} All models are fine-tuned exclusively with a next-token prediction objective. For fair comparison, hyperparameters are kept consistent across all experimental settings. We use the AdamW optimizer with a learning rate of 1e-6 and train for 1-2 epochs. For video processing, we uniformly sample 16 frames. 
% All experiments are conducted on 8 NVIDIA A100 GPUs and results are averaged over three independent runs.
See more details in Appendix~\ref{app:train_details}.

\textbf{Evaluation Metrics.} Our primary metric is \textit{Accuracy}, calculated via exact match with ground-truth answers. To assess generalization, we perform cross-dataset evaluation, where a model is trained on one dataset (e.g., NExT-QA) and tested on another unseen dataset (e.g., STAR).

\subsection{Synthetic Data Quality Assessment}
\begin{wrapfigure}{r}{0.5\textwidth} % 控制整体宽度
    \vspace{-4mm} % 根据需要调整整体上移
    \centering
    % 第一行：表格
    \begin{minipage}{\linewidth}
        \centering
        \captionof{table}{Statistics of the datasets we used.}\label{tab:raw_statistics} 
        % \resizebox{\linewidth}{!}{
            \begin{tabular}{c|cc}
            \hline
            & \#video & \#QA(Annotation)  \\
            \hline
            NExT-QA &3.8k & 34k  \\
            STAR &3k & 45k  \\
            DiDeMo & 2k &7k\\
            \hline
            \end{tabular}
            % }
    \end{minipage}
% \end{wrapfigure}
% \begin{wrapfigure}{r}{0.55\textwidth}
    \vspace{1mm} % 两行之间空隙
    % 第二行：两个图左右分列
    \begin{minipage}{0.48\linewidth}
        \includegraphics[width=\linewidth]{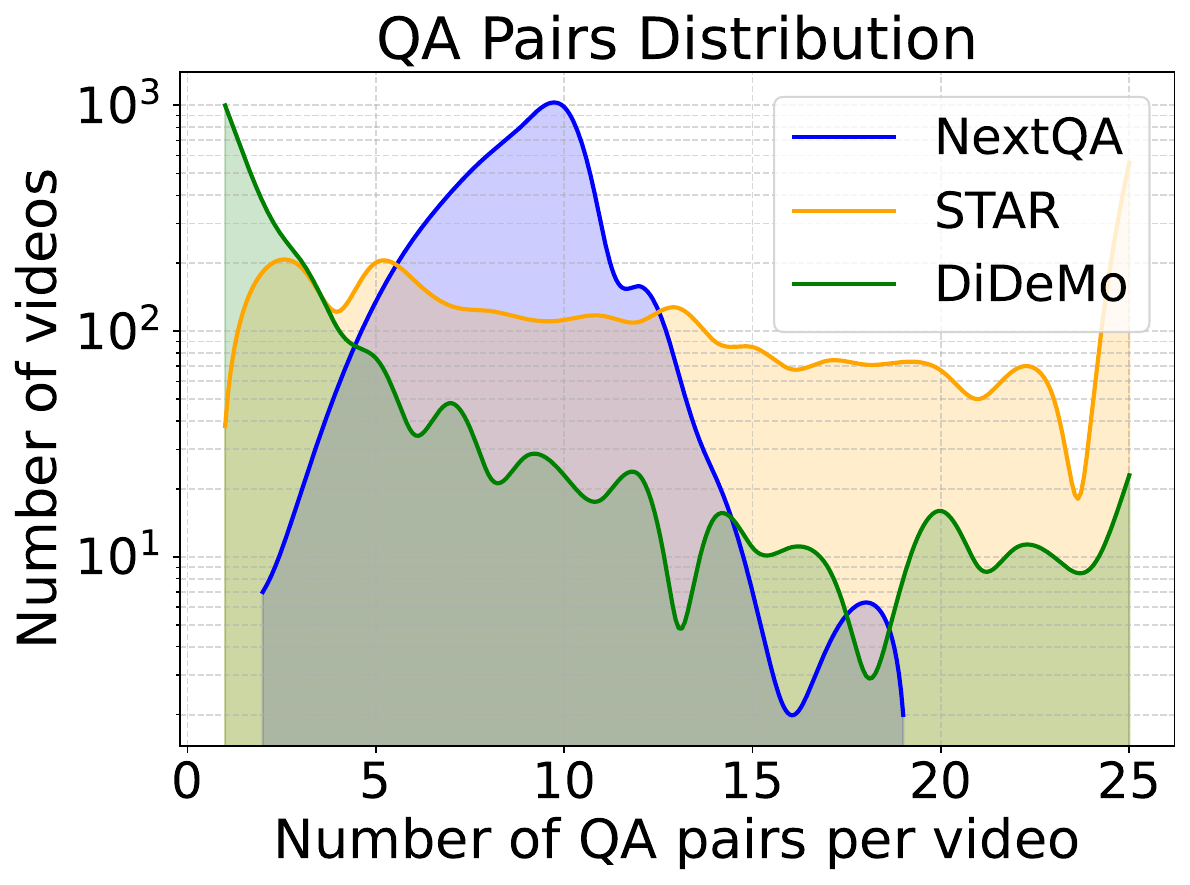}
        \caption{Distribution of the number of QA pairs per video across datasets. 
        NExT-QA and STAR include a wide range of annotations per video, 
        with some clips having more than 20 questions, while DiDeMo remains consistently sparse.}\label{fig:qa_dist}
    \end{minipage}
    \hfill
    \begin{minipage}{0.48\linewidth}
        \includegraphics[width=\linewidth]{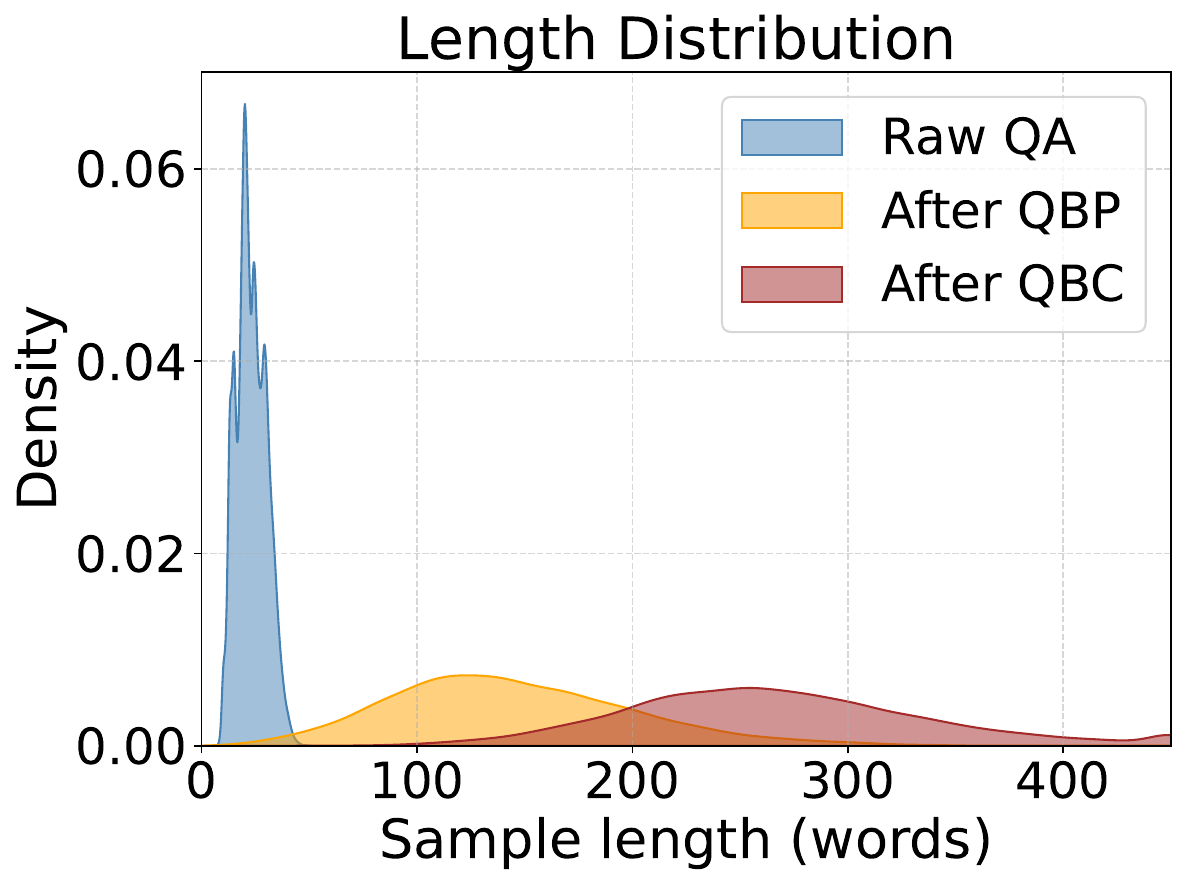}
        \caption{Length distributions of textual supervision before and after synthesis. 
        Raw QA pairs are short and fragmented. QBP generates moderately longer narratives 
        with high semantic density, while QBC produces the longest, fine-grained captions.}\label{fig:length}
    \end{minipage}
    \vspace{-4mm} % 让整体更紧凑
\end{wrapfigure}

\paragraph{Data Analysis.} Our data synthesis process begins with three widely-used VideoQA benchmarks as seeds: NExT-QA, STAR, and DiDeMo~\citep{xiao2021next,wu2021star,anne2017localizing}. As shown in Table~\ref{tab:raw_statistics}, these datasets exhibit notably different annotation densities. NExT-QA and STAR provide relatively dense supervision, with an average of 9 and 15 QA pairs per video, respectively. In contrast, DiDeMo is much sparser, with only 3.5 QA pairs per video. This disparity is further visualized in Figure~\ref{fig:qa_dist}. These statistics underscore the complementary nature of our proposed methods. The high density of questions in datasets like STAR provides a rich source for QBP to consolidate into coherent narratives. Conversely, the sparsity of datasets like DiDeMo highlights the need for QBC to expand annotation coverage with fine-grained, grounded descriptions.

Next, we analyze the textual properties of our synthesized data. Figure~\ref{fig:length} compares the length distributions. The original answers are predominantly short and fragmented. In contrast, QBP narratives are moderately longer and exhibit high semantic density, while QBC rationales produce the longest and most detailed descriptions. This analysis confirms that our framework successfully transforms sparse annotations into two distinct and complementary forms of supervision: one focused on semantic density (QBP) and the other on descriptive richness (QBC).

% \subsubsection{Assessing the Quality of Synthetic Data}

\paragraph{Quantitative Human Evaluation.} While our synthesis process is seeded with human-annotated QA pairs, the LLM or MLLM generator could potentially introduce errors. To rigorously evaluate this, we conduct a human evaluation study on the quality of our generated data. We randomly sample 100 QBP narratives and 100 QBC rationales and ask three human evaluators to rate them on a 1-5 Likert scale across several key dimensions. Instructions for human evaluators are in the App.~\ref{app:human_eval}.

% \begin{table}[t]
% \centering
% \caption{Human evaluation of synthetic data quality on a 1-5 scale. The scores are consistently high, confirming the overall quality and fidelity of the synthesized data.}
% \label{tab:human_eval}
% \resizebox{0.6\linewidth}{!}{
% \begin{tabular}{l|cc}
% \toprule
% \textbf{Evaluation Dimension} & \textbf{QBP } & \textbf{QBC } \\
% \midrule
% Factual Consistency & 4.21 $\pm$ 0.55 & 4.35 $\pm$ 0.48 \\
% Logical Coherence   & 4.25 $\pm$ 0.61 & - \\
% Visual Grounding    & -               & 4.38 $\pm$ 0.52 \\
% Fluency             & 4.88 $\pm$ 0.21 & 4.91 $\pm$ 0.19 \\
% \bottomrule
% \end{tabular}
% }
% \end{table}
\begin{wraptable}{r}{0.56\textwidth} % r=右侧, l=左侧, 0.4\textwidth=表格宽度
\vspace{-2mm}
\centering
\caption{Human evaluation of synthetic data quality on a 1-5 scale. The scores are consistently high, confirming the overall quality and fidelity of the synthesized data.}\label{tab:human_eval}
\resizebox{0.99\linewidth}{!}{
\begin{tabular}{l|cc}
\toprule
\textbf{Evaluation Dimension} & \textbf{QBP } & \textbf{QBC } \\
\midrule
Factual Consistency & 4.21 $\pm$ 0.55 & 4.35 $\pm$ 0.48 \\
Logical Coherence   & 4.25 $\pm$ 0.61 & - \\
Visual Grounding    & -               & 4.38 $\pm$ 0.52 \\
Fluency             & 4.88 $\pm$ 0.21 & 4.91 $\pm$ 0.19 \\
\bottomrule
\end{tabular}
}
% \vspace{-7mm}
\end{wraptable}

As shown in Table~\ref{tab:human_eval}, our synthesized data achieves consistently high scores. Both QBP and QBC demonstrate strong \textit{Factual Consistency} (4.21 and 4.35, respectively), confirming that the LLM and MLLM generally preserve the ground-truth information from the source QA pairs. QBP narratives are rated favorably for \textit{Logical Coherence} (4.25), while QBC rationales receive a high score for \textit{Visual Grounding }(4.38). The low standard deviation across most metrics, particularly \textit{Fluency}, indicates strong agreement among evaluators on the high quality of the generated text. These results confirm that our synthesis process produces reliable supervision signals suitable for model training.

% --- Insert the Human Evaluation Table Here ---

\paragraph{Qualitative Error Analysis.}~\label{sec:error}
We perform a qualitative analysis of failure cases to better understand the limitations of our approach. We find that severe errors, such as hallucinating non-existent events, are extremely rare. The more common, though still infrequent, failure modes are subtle and method-specific. For QBP, the primary challenge lies in logical cohesion. We observe occasional errors in temporal ordering, where the LLM incorrectly sequences two closely related actions. This issue is sometimes exacerbated by imprecise temporal boundary annotations in the source QA pairs themselves, which provide ambiguous cues. In rarer cases, we note entity confusion, where a single person is described with conflicting pronouns as if they were two separate individuals. For QBC, the most notable failure mode is a form of "justified fabrication." Since the MLLM is provided with the correct answer, it sometimes invents plausible-sounding visual details to rationalize the answer, especially when the actual visual evidence is subtle or ambiguous. 
Detailed examples are provided in the Appendix~\ref{app:examples}.
% \ljx{See Appendix~\ref{app:assess} for more details}. 
% \ljxc{maybe 3 lines space left in page 9?}
% it sometimes invents plausible-sounding visual details to rationalize the answer, especially when the actual visual evidence is subtle or ambiguous.
% it sometimes invents plausible-sounding visual details to rationalize the answer, especially when the actual visual evidence is subtle or ambiguous.it sometimes invents plausible-sounding visual details to rationalize the answer, especially when the actual visual evidence is subtle or ambiguous.

\subsection{Main Performance Evaluation}

We evaluate our data synthesis framework by comparing it against previously published state-of-the-art (SOTA) models (e.g., Vamos~\citep{wang2023vamos}, MotionEpic~\citep{fei2024video}), which are fine-tuned on the original QA training sets of each benchmark. Further details on these baselines are provided in Appendix~\ref{app:baselines}. For our evaluation, we select two representative MLLMs, Qwen2.5-VL~\citep{bai2025qwen25vltechnicalreport} and MiMo-VL~\citep{coreteam2025mimovltechnicalreport}, chosen for their strong general-purpose reasoning ability. Table~\ref{tb:main_results} summarizes the results on NExT-QA and STAR.
% It is worth noting that these models are pretrained on large-scale web data, which may contain samples overlapping with common benchmarks as shown in~\cite{li2024llava}.
% Table~\ref{tb:main_results} summarizes the results on NExT-QA and STAR. The upper block reports the performance of prior SOTA models, while the lower block presents our core experiments.

The results are clear and consistent: across all backbones and model scales, training exclusively on our synthesized data provides significant performance improvements. For example, when applied to the Qwen2.5-VL-3B, our method boosts accuracy on NExT-QA from 74.3\% to 76.8\% (+2.5) and delivers a remarkable +5.0 point gain on STAR, increasing accuracy from 67.5\% to 72.5\%. This trend holds for larger 7B models as well; notably, our method pushes the Qwen2.5-VL-7B model to a new SOTA of 80.8\% on NExT-QA. Importantly, even when built upon strong backbones, our synthesized supervision consistently outperforms models fine-tuned on the raw QA pairs, underscoring its effectiveness and general applicability.

% Table~\ref{tb:main_results} summarizes the results on NExT-QA and STAR. The results demonstrate a significant performance uplift, even on top of these strong, pre-exposed baselines. For instance, while the baseline Qwen2.5-VL-7B already achieves a very high 76.2\% on NExT-QA (likely benefiting from its pre-training), our synthetic supervision pushes this performance further to a new state-of-the-art of **80.8\% (+4.6 points)**. This pattern is consistent across all tested backbones. On STAR, where models often struggle more, our method boosts the Qwen2.5-VL-3B from 67.5\% to **72.5\%**, a substantial +5.0 point gain.

% This finding is particularly significant. It suggests that even when a model has already been trained on the original QA pairs in their raw, fragmented format, its potential remains constrained. 
% By reorganizing and enriching the supervision signal into coherent narratives (QBP) and grounded rationales (QBC), our method allows the model to build a deeper, more structured understanding of event logic and visual evidence. This ability to further enhance already powerful and heavily pre-trained models underscores the critical value of improving the \textit{form} and \textit{structure} of supervision, moving beyond the limitations of the "bag-of-facts" paradigm.

This consistent improvement validates our core hypothesis: transforming the supervision format from a "bag-of-facts" into structured, multi-level signals is a more effective way to train VideoQA models. The narrative-level context from QBP enables models to better understand temporal and causal event structures, while the instance-level grounding from QBC forces a tighter alignment between reasoning and specific visual evidence. This richer supervision allows models to move beyond shallow pattern matching and develop a deeper, more robust comprehension of video content, leading to higher accuracy on complex reasoning tasks.

\begin{table}[t]
\centering
\caption{Model comparison on NExT-QA and STAR. All scores are reported in Accuracy (\%).}\label{tb:main_results}
% \resizebox{\linewidth}{!}{
\begin{tabular}{c|c|c|c}
\toprule
\multirow{1}{*}{Model}  &\multirow{1}{*}{LLM Arch.} &\multicolumn{1}{c|}{NExT-QA}  &\multicolumn{1}{c}{STAR}  
\\
 \midrule
 \multicolumn{4}{l}{\textit{Fine-tuned on Raw QA pairs}} \\
InternVideo~\citep{wang2022internvideo} &  -     &63.2   &58.7      \\

% \color{gray}
% LSTP~\citep{wang2024lstp} &\color{gray}FlanT5\ 3B   &\color{gray}72.1    &\color{gray}-    \\
% % \color{gray} 
% SeViLA~\citep{yu2023self} &\color{gray}FlanT5\ 3B   &\color{gray}73.8 &\color{gray}67.2   \\ 
% % \color{gray} 
% VidF4~\citep{liang2024end} &\color{gray}FlanT5\ 3B    & \color{gray}74.1  & \color{gray}68.1  
% \\ 
LSTP~\citep{wang2024lstp} &FlanT5\ 3B   &72.1    &-    \\
%  
% SeViLA~\citep{yu2023self} &FlanT5\ 3B   &73.8 &67.2   \\ 
%  
VidF4~\citep{liang2024end} &FlanT5\ 3B    & 74.1  & 68.1  
\\ 
LLaMA-VQA~\citep{ko2023large} &LLaMA\ 7B  &72.0 
&65.4    \\
% \rowcolor{green!2}\color{gray}
MotionEpic~\citep{fei2024video} & Vicuna\ 7B    &76.0 
  &71.0   \\ 
Vamos~\citep{wang2023vamos} & LLaMA2\ 7B   &75.0 
&-   \\
LLaVA-OV\citep{li2024llava} & Qwen2\ 7B   &77.5 &66.2   \\
\midrule
% \color{gray}
% Qwen2.5-VL~\citep{bai2025qwen25vltechnicalreport} & \color{gray}Qwen2.5\ 3B    &\color{gray}74.3 &\color{gray}67.5   \\ 
% w. QBP+QBC\ (ours) & Qwen2.5-3B   
% &\textbf{76.8}  &	\textbf{72.5} \\
 \multicolumn{4}{l}{\textit{Combined with on QBP+QBC (ours)}} \\
Qwen2.5-VL~\citep{bai2025qwen25vltechnicalreport} & Qwen2.5\ 3B    &74.3 &67.5   \\ 
w. QBP+QBC\ (ours) & Qwen2.5-3B   
&76.8~\color{blue}(+2.5)  &	72.5~\color{blue}(+5.0) \\

% \color{gray}
% MiMo-VL-SFT~\citep{coreteam2025mimovltechnicalreport} & \color{gray}MiMo\ 7B    &\color{gray}75.3 
%   &\color{gray}52.0   \\ 
MiMo-VL-SFT~\citep{coreteam2025mimovltechnicalreport} & MiMo\ 7B    &75.3 
  &52.0   \\ 
  
w. QBP+QBC\ (ours)  & MiMo\ 7B   
&77.0~\color{blue}(+1.7)  &	56.2~\color{blue}(+4.2) \\
% \color{gray}
% Qwen2.5-VL~\citep{bai2025qwen25vltechnicalreport} & \color{gray}Qwen2.5\ 7B    &\color{gray}76.2 &\color{gray}70.6   \\ 
Qwen2.5-VL~\citep{bai2025qwen25vltechnicalreport} &Qwen2.5\ 7B    &76.2 &70.6   \\ 
w. QBP+QBC\ (ours) & Qwen2.5\ 7B    &80.8~\color{blue}(+4.6) 
  &73.3~\color{blue}(+2.7)   \\
\bottomrule
\end{tabular}
% }
\vspace{-3mm}
\end{table}

\subsection{In-depth Analysis of Question-Based Paraphrasing (QBP)}

In this section, we conduct a detailed analysis to understand the properties of QBP. We aim to answer two key questions: (1) How does fine-tuning on QBP-synthesized narratives compare to fine-tuning on raw QA pairs, particularly concerning cross-dataset generalization? (2) How does the choice of seed dataset for synthesis affect QBP's performance?

% \subsubsection{QBP Mitigates Overfitting from Raw Data}

\paragraph{QBP Mitigates Overfitting from Raw Data.} A common risk in fine-tuning is overfitting to the source dataset's specific patterns and biases. To investigate whether QBP can mitigate this issue, we compare models fine-tuned on raw QA pairs from a single source against models fine-tuned on QBP narratives synthesized from that same source.

Table~\ref{tab:qbp_vs_raw} reveals a critical trend. As expected, fine-tuning the backbone on the raw NExT-QA training set improves its in-domain performance significantly (+1.9\%), but this comes at the cost of degraded performance on the unseen STAR dataset (-1.1\%), a clear sign of overfitting. Conversely, training on QBP narratives synthesized from NExT-QA not only boosts in-domain accuracy but also enhances cross-dataset generalization to STAR (+2.2\%). The same pattern holds when using STAR as the source dataset. This directly validates that the narrative supervision from QBP provides a more generalizable signal than the original, fragmented QA pairs.

\paragraph{Effect of Diverse Seeds for QBP Synthesis.} Next, we explore how leveraging a diverse mix of seed datasets for QBP synthesis impacts performance. We generate QBP narratives using various combinations of NExT-QA, STAR, and DiDeMo as source material.
As shown in Table~\ref{tab:qbp_seeds}, combining seeds from multiple datasets yields the most significant gains, particularly for cross-domain generalization. While narratives from a single source already provide benefits, synthesizing from a mix of NExT-QA and STAR pushes the STAR accuracy to a high of 70.9\% (+3.3\% over the backbone). Incorporating all three diverse sources (NExT-QA, STAR, DiDeMo) achieves the best overall balance, reaching 76.5\% on NExT-QA and 70.8\% on STAR. This confirms that QBP is most effective when it can draw upon a wide range of question styles and content, allowing it to generate a richer and more robust narrative supervision signal that transcends the biases of any single dataset.

\begin{table}[t]
\centering
\begin{minipage}{0.46\textwidth} % 左表占比
\centering
\caption{Comparison of fine-tuning on raw QA pairs vs. our QBP-synthesized narratives. QBP effectively improves in-domain performance while also enhancing cross-domain generalization, mitigating the overfitting seen with raw data.}\label{tab:qbp_vs_raw}
\resizebox{\linewidth}{!}{ % 自动缩放到 minipage 宽度
\begin{tabular}{l|cc}
\toprule
\textbf{Training Data} & \textbf{Test on NExT-QA} & \textbf{Test on STAR} \\
\midrule
\textit{Qwen2.5-VL-3B} & \textit{74.3} & \textit{67.6} \\
\midrule
% Fine-tuned on NExT-QA (raw) & 76.2 (+1.9) & 66.5 {\color{red}(-1.1)} \\
% Fine-tuned on \textbf{QBP from NExT-QA} & 76.0 (+1.7) & 69.8 {\color{blue}(+2.2)} \\
% \midrule
% Fine-tuned on STAR (raw) & 73.1 {\color{red}(-1.2)} & 70.2 ((+2.6) \\
% Fine-tuned on \textbf{QBP from STAR} & 75.5 {\color{blue}(+1.2)} & 69.9 ((+2.3) \\
NExT-QA (raw) & 76.2 (+1.9) & 66.5 {\color{red}(-1.1)} \\
\textbf{QBP from NExT-QA} & 76.0 (+1.7) & 69.8 {\color{blue}(+2.2)} \\
\midrule
STAR (raw) & 73.1 {\color{red}(-1.2)} & 70.2 ((+2.6) \\
\textbf{QBP from STAR} & 75.5 {\color{blue}(+1.2)} & 69.9 ((+2.3) \\

\bottomrule
\end{tabular}
}
\end{minipage}%
\hspace{1mm} % 两表间距
\begin{minipage}{0.49\textwidth} % 右表占比
\centering
\caption{Effect of using different and combined seed datasets for QBP synthesis. Performance is evaluated on NExT-QA and STAR. Combining diverse seeds yields the best generalization.}\label{tab:qbp_seeds}
\resizebox{\linewidth}{!}{
\begin{tabular}{l|cc}
\toprule
\textbf{QBP Seed Data Source(s)} & \textbf{Test on NExT-QA} & \textbf{Test on STAR} \\
\midrule
\textit{Qwen2.5-VL-3B (No fine-tuning)} & \textit{74.3} & \textit{67.6} \\
\midrule
NExT-QA only & 76.0 & 69.8 \\
DiDeMo only & 76.0 & 69.1 \\
STAR only & 75.5 & 69.9 \\
\midrule
NExT-QA + DiDeMo & 76.3 & 69.4 \\
NExT-QA + STAR & 76.2 & \textbf{70.9} \\
NExT-QA + DiDeMo + STAR & \textbf{76.5} & 70.8 \\
\bottomrule
\end{tabular}
}
\end{minipage}
\end{table}
% --- TABLE 1: QBP vs. Raw Fine-tuning ---
% \input{2_table_fig/qbp_vs_raw}
% --- TABLE 2: Effect of Diverse QBP Seeds ---
% \input{2_table_fig/qbp_seeds}
% \subsubsection{Effect of Diverse Seeds for QBP Synthesis}
% --- FIGURE 3: QBC SCALING  ---

% \begin{figure}
% \centering
%   \begin{minipage}[t]{0.497\linewidth} 
%     \centering
%     \subfigure{
%     \includegraphics[width=\textwidth]{1_pic/accuracy_curve_nextqa.pdf}}
%     % \caption{Impact of using only single.}\label{fig:kthlast}
%     % \vspace{-3mm}
%   \end{minipage} 
%   \begin{minipage}[t]{0.497\linewidth} 
%     \centering
%     \subfigure{
%     \includegraphics[width=\textwidth]{1_pic/accuracy_curve_star.pdf}}
%     % \caption{Impact of using only e.}\label{fig:kthlast}
%     % \vspace{-3mm}
%   \end{minipage} 
%   % \begin{minipage}[t]{0.32\linewidth} 
%   %   \centering
%   %   \subfigure[E-commerce]{
%   %   \includegraphics[width=0.95\textwidth]{ACL/picture/baseline ecd.pdf}}
%   %   % \caption{Impact of using only single utterance close to response.}\label{fig:kthlast}
%   %   % \vspace{-3mm}
%   % \end{minipage} 
%   \caption{Effect of QBC scale. Larger amounts of synthesized QBC data improve accuracy and convergence on both NExT-QA and STAR, with clear gains in cross-dataset generalization.}\label{fig:compare}
% \end{figure}
\begin{figure}[!t]
\vspace{-4mm}
    \centering
    % 左侧图，占50%宽度，顶部对齐
    \begin{minipage}[c]{0.65\linewidth} 
    \centering
    \subfigure{
    \includegraphics[width=0.483\textwidth]{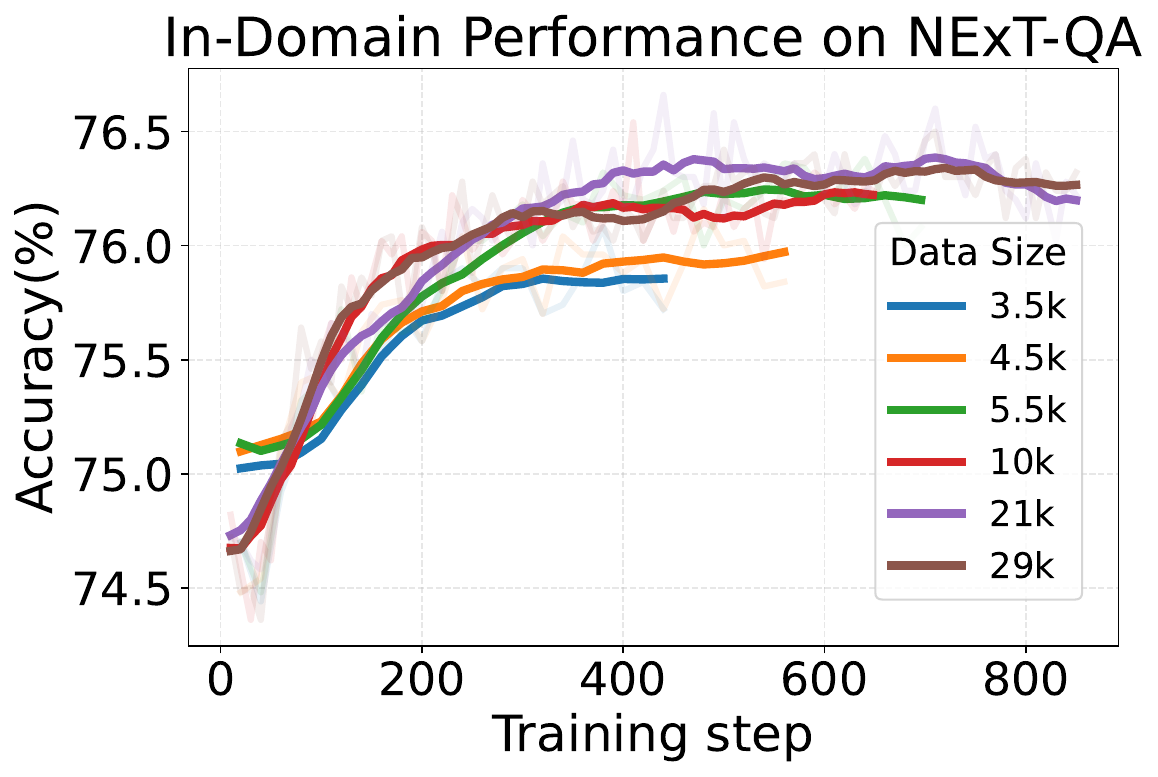}}
    \hfill
    \subfigure{
    \includegraphics[width=0.483\textwidth]{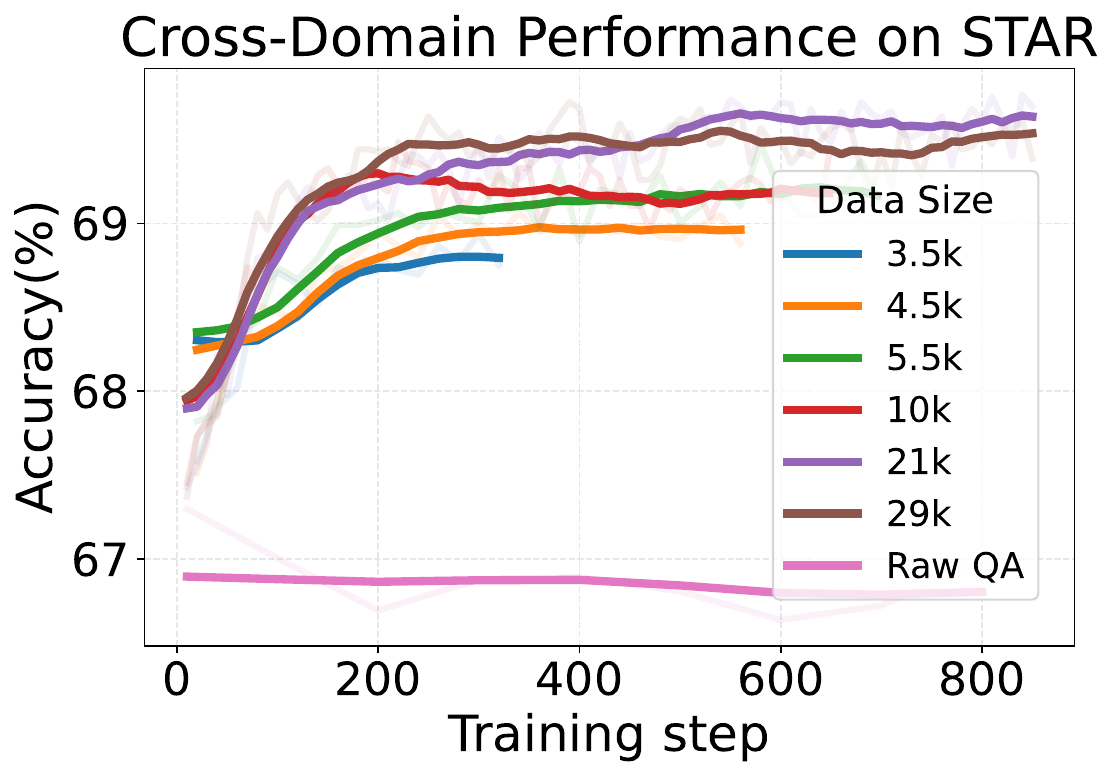}}
    \caption{Effect of QBC scale. Larger amounts of synthesized QBC data improve accuracy and convergence on both NExT-QA and STAR, with clear gains in cross-dataset generalization. }
    \label{fig:qbc-scale}
    \end{minipage}
    % \hfill
    % 右侧图，占50%宽度，顶部对齐    
    % \hfill
    \begin{minipage}[c]{0.01\textwidth}
        \centering
        \begin{tikzpicture}
            \draw[dashed] (0,0.) -- (0,5.);  % 竖直虚线，长度为5个单位
        \end{tikzpicture}
    \end{minipage}
    % \hfill
    \begin{minipage}[c]{0.28\linewidth} 
    \centering
    \subfigure{
    \includegraphics[width=\textwidth]{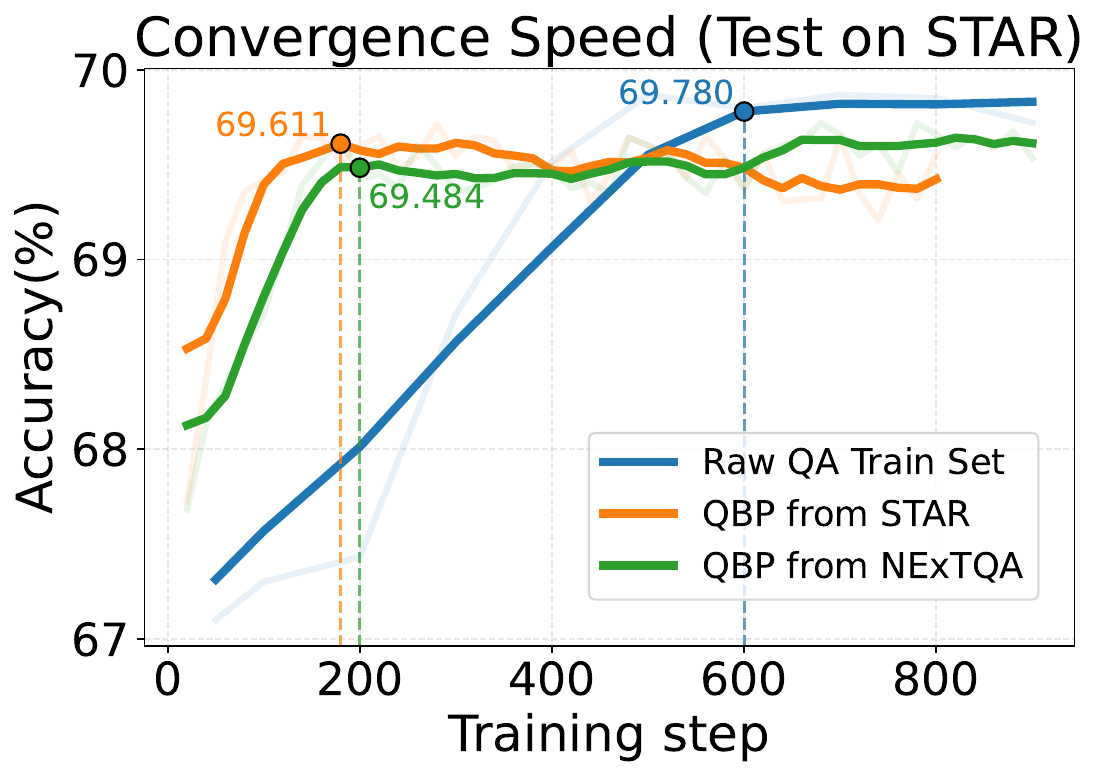}}
    \caption{Convergence with raw QA vs. QBP. QBP enables faster convergence compared to raw QA training, showing the efficiency of holistic descriptions.}
    \label{fig:convergence}
    % \vspace{-3mm}
  \end{minipage} 
        
    % \end{subfigure}

\end{figure}

% \subsubsection{Accelerated Convergence with QBP's Narrative Supervision}

\paragraph{Accelerated Convergence with QBP's Narrative Supervision.} A striking finding of our study concerns the training efficiency of QBP. As shown in Figure~\ref{fig:convergence}, models trained on QBP-synthesized narratives converge dramatically faster than those trained on the original, fragmented QA pairs. For instance, the NExT-QA training set consists of approximately 30k individual QA pairs, which our QBP process condenses into about 3k holistic paragraphs. Despite this ten-fold reduction in the number of training instances, the model trained on QBP data reaches its performance plateau within approximately 220 steps. In stark contrast, the model trained on the raw QA set requires around 600 steps to reach a similar performance level. This demonstrates that our QBP-based supervision accelerates convergence by more than 2.5x compared to the standard QA training paradigm.

This accelerated convergence may seem counterintuitive, given that QBP narratives are textually longer than individual QA pairs. We hypothesize this is because the narrative supervision acts as a far more semantically dense and informative training signal. Each paragraph synthesizes multiple related inquiries (`what`, `how`, `why`) and their dependencies into a unified context that reflects the video's underlying event structure (see Table~\ref{tab:intro_example}). This provides the model with richer, more structured reasoning cues in a single optimization step, effectively reducing the redundancy inherent in processing numerous overlapping, low-level QA pairs.

From a practical standpoint, this highlights a significant efficiency advantage of QBP. In resource-constrained settings, the ability to reach a high-performance state with fewer training steps makes QBP-based supervision a particularly appealing and cost-effective strategy.

% For QBC, we perform a controlled study by varying the amount of generated rationales used in training. We find that model performance scales positively with the volume of QBC-generated data, indicating that visual rationales provide a strong and scalable training signal. Furthermore, cross-dataset experiments show that QBC improves generalization: models trained on QBC-enhanced NExT-QA data achieve $+3$–$4\%$ higher accuracy on STAR compared to models trained on the original NExT-QA training set. This suggests that QBC promotes more transferable reasoning, as the synthesized rationales encourage grounding in visual evidence rather than dataset-specific biases.

% This section follows the "In-depth Analysis of QBP" subsection

\subsection{Analysis of QBC: Data Scaling and Generalization}

Having analyzed QBP's properties with respect to seed diversity, we now turn to QBC and investigate its effectiveness as a function of data scale. We synthesize varying amounts of QBC rationales from the NExT-QA training set (from 3.5k up to the full 29k samples) and fine-tune the Qwen2.5-VL-3B backbone on each subset. Performance is monitored on both the in-domain (NExT-QA) and cross-domain (STAR) test sets.

The results, plotted in Figure~\ref{fig:qbc-scale}, show a clear and positive correlation between the volume of synthetic data and model performance. 
% \begin{itemize}
%     \item \textbf{In-domain Performance (Fig.~\ref{fig:qbc-scale}a):} On NExT-QA, accuracy steadily improves as more QBC rationales are added. With just 5k samples, the model already surpasses the baseline, and performance continues to climb as the dataset scales to 10k and then 29k samples. This confirms that the fine-grained, grounded supervision provided by QBC offers a strong and scalable training signal for improving in-domain reasoning.
%     \item \textbf{Cross-dataset Generalization (Fig.~\ref{fig:qbc-scale}b):} The benefits of scaling QBC data are even more pronounced in the cross-dataset transfer setting. On STAR, performance rises from a baseline of 66.5\% with 0.5k samples to nearly 70.0\% with the full 29k set, a gain of +3.5 points. This striking trend demonstrates that training on QBC's visual rationales effectively forces the model to ground its predictions in visual evidence rather than source-specific linguistic biases, thereby enhancing its ability to generalize to new, unseen domains.
% \end{itemize}
(1)~\textbf{In-domain Performance (Fig.~\ref{fig:qbc-scale}a):} On NExT-QA, accuracy steadily improves as more QBC rationales are added. With just 5k samples, the model already surpasses the baseline, and performance continues to climb as the dataset scales to 10k and then 29k samples. This confirms that the fine-grained, grounded supervision provided by QBC offers a strong and scalable training signal for improving in-domain reasoning.
(2)~\textbf{Cross-dataset Generalization (Fig.~\ref{fig:qbc-scale}b):} The benefits of scaling QBC data are even more pronounced in the cross-dataset transfer setting. On STAR, performance rises from a baseline of 66.5\% (\textit{Raw QA}) to nearly 70.0\% with the full 29k set, a gain of +3.5 points. This striking trend demonstrates that training on QBC's visual rationales effectively forces the model to ground its predictions in visual evidence rather than source-specific linguistic biases, thereby enhancing its ability to generalize to new, unseen domains.

In summary, this analysis validates QBC as a highly scalable form of supervision. Increasing the volume of QBC data consistently improves both in-domain accuracy and, critically, cross-dataset generalization. This highlights its role as a powerful tool for generating fine-grained, evidence-based supervision that complements the holistic, narrative context provided by QBP.

\section{Conclusion}

In this work, we present a novel data-centric paradigm for VideoQA that moves beyond the limitations of training on isolated, factual annotations. Our framework introduces two complementary synthesis strategies: Question-based Paraphrasing (QBP), which generates coherent, narrative-level supervision, and Question-based Captioning (QBC), which provides fine-grained, instance-level visual grounding. Our extensive experiments demonstrate that training models exclusively on this synthesized data establishes a new state-of-the-art on multiple challenging benchmarks. Beyond accuracy, we show that our method yields significant secondary benefits: it substantially enhances cross-dataset generalization, and the narrative supervision from QBP markedly accelerates model convergence by more than 2.5x. Our rigorous human evaluation further confirms the high factual consistency and logical coherence of the synthesized data, solidifying its reliability as a high-quality supervision signal. 
These results highlight the profound potential of shifting focus from model architecture to the supervision signal itself. 
By transforming fragmented inquiries into structured narratives and grounded rationales, we unlock significant gains in model performance, robustness, and training efficiency. 

\section*{Ethics Statement}

This work builds on publicly available VideoQA datasets, which contain human-annotated QA pairs and captions. No private or sensitive data are used. Our data synthesis strategies (QBP and QBC) rely on large language models and multimodal models to generate additional supervision, but the generated content remains constrained to the semantics of the original annotations, reducing risks of misinformation or harmful outputs. Potential societal risks include over-reliance on synthetic data or propagation of biases from source models; we mitigate this by grounding synthesis in human-verified annotations and reporting transparent analyses. All experiments follow standard academic use of benchmarks and are intended solely for advancing research in multimodal reasoning.

\section*{Reproducibility Statement}
We provide detailed descriptions of our methods, datasets, and experimental settings to ensure reproducibility. Specifically, we outline the backbone architectures, frame sampling strategy, training objectives, and hyperparameters. Dataset splits follow publicly available benchmarks. Prompts used for QBP and QBC synthesis are included in the Appendix. We also report results averaged across multiple random seeds to account for variance. All resources required to reproduce our results including code, and processed data will be released upon publication.

% \section*{Limitations}
% Our method is straightforward but relies on the quality of reasoning generated by MLLMs. Although the performance of the VideoQA model improves significantly after refinement, the incorrect or biased reasoning produced by these MLLMs can still negatively affect the overall performance. In future work, we will continue to investigate how to generate higher-quality reasoning or develop better refinement strategies to enable the model to extract more consistent and meaningful information from the reasoning data.

% \section*{Acknowledgements}

% We would like to express our sincere gratitude to the anonymous reviewers for their thorough review,
% insightful comments, and constructive suggestions, which have significantly improved the quality
% of this manuscript. 
% % We also appreciate the valuable guidance and support provided by the AI Data Technology Laboratory of Huawei Noah's Ark Lab.
% This work paper is supported by the State Key Laboratory of General Artificial Intelligence.

\bibliography{iclr2026_conference}
\bibliographystyle{iclr2026_conference}

\appendix

\section{Use of Large Language Models}
arge language models are used in two ways in this work. First, they support data synthesis, where QBP relies on language models to paraphrase human-annotated QA pairs into narrative form, and QBC employs multimodal models to generate query-conditioned captions. Second, they play a supportive role in writing, including proofreading, correcting grammatical errors, and improving clarity of exposition. All model outputs are carefully reviewed by the authors, and responsibility for the final content rests entirely with the authors.

% \section{Experiments Setup}
\section{Experimental Details}

\paragraph{Training details}\label{app:train_details}
We finetune model using the SFTTrainer from TRL~\footnote{\url{https://huggingface.co/docs/trl/v0.22.1/en/sft_trainer}} and DeepSpeed~\footnote{\url{https://github.com/deepspeedai/DeepSpeed}} during training in NVIDIA H800 (80GB) GPU $\times$ 2. We use AdamW with a cosine learning rate scheduler, whose max learning rate is 1e-6, and a batch size of 8. We train our model within 1-2 epochs. Our training code will be later open sourced.
% Our training code is implemented based on LAVIS~\footnote{\url{https://github.com/salesforce/LAVIS}} and transformers~\footnote{\url{https://github.com/huggingface/transformers}} libraries, and will be later open sourced.

% \textbf{BLIP-2} processes all frames by voting or concatenating them and then uses LLM to generate the final answer.

\paragraph{Baselines.}\label{app:baselines} We evaluate our data synthesis framework by comparing it against previously SOTA models, such as InternVideo \citep{wang2022internvideo}, LLaMA-VQA \citep{ko2023large}, LSTP \citep{wang2024lstp}, VidF4~\citep{liang2024end}, Vamos~\citep{wang2023vamos}, MotionEpic~\citep{fei2024video} and LLaVA-OV~\cite{li2024llava}. Among these models, LLaMA-VQA, Vamos, and MotionEpic use 7B-parameter LLM as part of the model.

\textbf{LSTP} adopts the BLIP-2 architecture and uses optical flow for frame selection, followed by using LLM to generate answers. Similarly, VidF4~\citep{liang2024end} update its model by training on raw QA pairs after extracting key frames from videos.

\textbf{LLaMA-VQA} is built based on LLaMA-7B~\citep{touvron2023llama}, enabling the model to understand the complex relationships between videos, questions, and answers by constructing multiple auxiliary tasks.

% \textbf{SeVILA} relies on a multi-stage training process and is trained on an additional dataset with temporal localization supervision. During the inference, it first utilizes BLIP-2 and LLMs for frame selection and then uses BLIP-2 and LLM again for answer generation. 

\textbf{MotionEpic} breaks down the raw intricate video reasoning problem into a chain of simpler sub-problems and solves them one by one sequentially.

\textbf{Vamos}~\citep{wang2023vamos} generalizes the concept bottleneck model to work with tokens and nonlinear models, which uses hard attention to select a small subset of tokens from the free-form text as inputs to the LLM reasoner. 

\textbf{LLaVA-OV}~\citep{li2024llava} builds upon LLaVA by constructing synthetic data to further enhance the base model. \cite{liang-etal-2025-reasvqa} fine-tune the model on raw QA pairs for VideoQA tasks. However, the official paper also shows that current MLLM backbones may overlap with common benchmarks; see the original work for details.

\section{Prompts and Examples for QBP and QBC}

\subsection{Prompts and Examples for Data Synthesis}

Here, we provide the detailed prompts and concrete examples used for our data synthesis strategies.

\subsubsection{Question-based Captioning (QBC)}

The QBC prompt instructs the MLLM to generate a visual rationale—a caption that describes the visual evidence supporting a given answer, without explicitly stating the answer itself.

\begin{promptbox}{Prompt for QBC}
% --- CHANGE START ---
% Same change as above: remove 'verbatim', use 'texttt' and '\\'.
\texttt{
Given a video, a question, and its answer, generate a natural language caption that highlights the visual content most relevant to justifying the answer. \\
\\
The caption should be a descriptive proof grounded in visual evidence, NOT a direct restatement of the answer.
}
% --- CHANGE END ---
\end{promptbox}

% \paragraph{Example.}
% We use an example related to the same snowmobile scene.

% \begin{promptbox}{Input for QBC}
% \begin{itemize}
%     \item \textbf{Frame:} A man in red is sitting on a snowmobile next to other people in a snowy landscape.
%     \item \textbf{Question:} Why is the person in red sitting on a snowmobile with a group?
%     \item \textbf{Answer:} resting
% \end{itemize}
% \end{promptbox}

% \begin{promptbox}{Generated QBC Visual Rationale}
% \begin{quote}
% Several people are gathered together with their parked snowmobiles in a snowy landscape. The person in red is seated calmly on one of the vehicles while others wait nearby, suggesting a pause in their activity.
% \end{quote}
% \end{promptbox}

\subsubsection{Question-based Paraphrasing (QBP)}

The QBP prompt is designed to instruct the LLM (DeepSeek, GPT-4o) to act as a reasoning integrator, synthesizing a holistic narrative from a collection of fragmented QA pairs.

\begin{promptbox}{Prompt for QBP}
% --- CHANGE START ---
% We remove the 'verbatim' environment and use 'texttt' for the font.
% Long lines will now wrap automatically. We use '\\' for manual line breaks.
\texttt{Transform the following Q\&A pairs into a single, logically coherent paragraph in present tense. Follow these rules strictly: \\
\\
1. **Content Requirements**: \\
- Use ONLY information from the provided Q\&A pairs. Do not invent new facts.\\
- If questions imply clear chronological order (e.g., "before"/"after"), preserve it.\\
       - If no temporal relationship exists (e.g., between weather and object questions), present facts in neutral order without implying sequence (avoid "first"/"then"/"while"). \\
       - Group related facts by theme (e.g., environment → actions → social interactions).\\
    2. **Prohibitions**:\\
       - Never assume unstated temporal/causal relationships.\\
       - Avoid speculative terms like "probably", "might", or "seems".\\
       - Do not use filler phrases like "the questions ask about...".\\
\\
    **Q\&A Pairs**:\\
    \{QA\ Group\} \\
\\
    **Output**:\\
}
% --- CHANGE END ---
\end{promptbox}

\paragraph{Example.}
To illustrate QBP's ability to reconstruct a complex event structure, we use the same set of questions previously shown in Table~\ref{tab:intro_example} of the introduction.

\begin{promptbox}{Input QA Pairs for QBP (from Table~\ref{tab:intro_example})}
\begin{itemize}
    \item Q1: How are the people transported on snow? (snowmobile)
    \item Q2: What is the weather like? (cold)
    \item Q3: Why is the person in red sitting on a snowmobile? (resting)
    \item Q4: How does the man in black react to the camera? (poses)
    \item Q5: Why have the snowmobiles parked? (resting)
    \item Q6: What is the relationship between the people? (friends)
    \item Q7: Why is the man in blue holding a camera? (to take a photo)
    \item Q8: What does the man in red do? (takes a photo)
\end{itemize}
\end{promptbox}

\begin{promptbox}{Synthesized QBP Narrative}
\begin{quote}
In cold weather conditions, a group of friends is transported on a snowmobile across the snow. They come to a halt and park the snowmobile to rest. The man wearing red, after placing his helmet on the motorbike, takes a photo. Meanwhile, the person dressed in red sits on the snowmobile, resting alongside the group. The man in black strikes a pose when the man in red raises his camera at the end of the video. In the group, there is also a man wearing blue who holds a camera to take a photo.
\end{quote}
\end{promptbox}

This example demonstrates how QBP synthesizes descriptive (`cold`, `snowmobile`), causal (`resting`), and social (`friends`, `poses`) information from the fragmented QA pairs into a single, logic-infused narrative.

\subsection{Examples}\label{app:examples}
\begin{figure}[t]
    \centering
    \includegraphics[width=\linewidth]{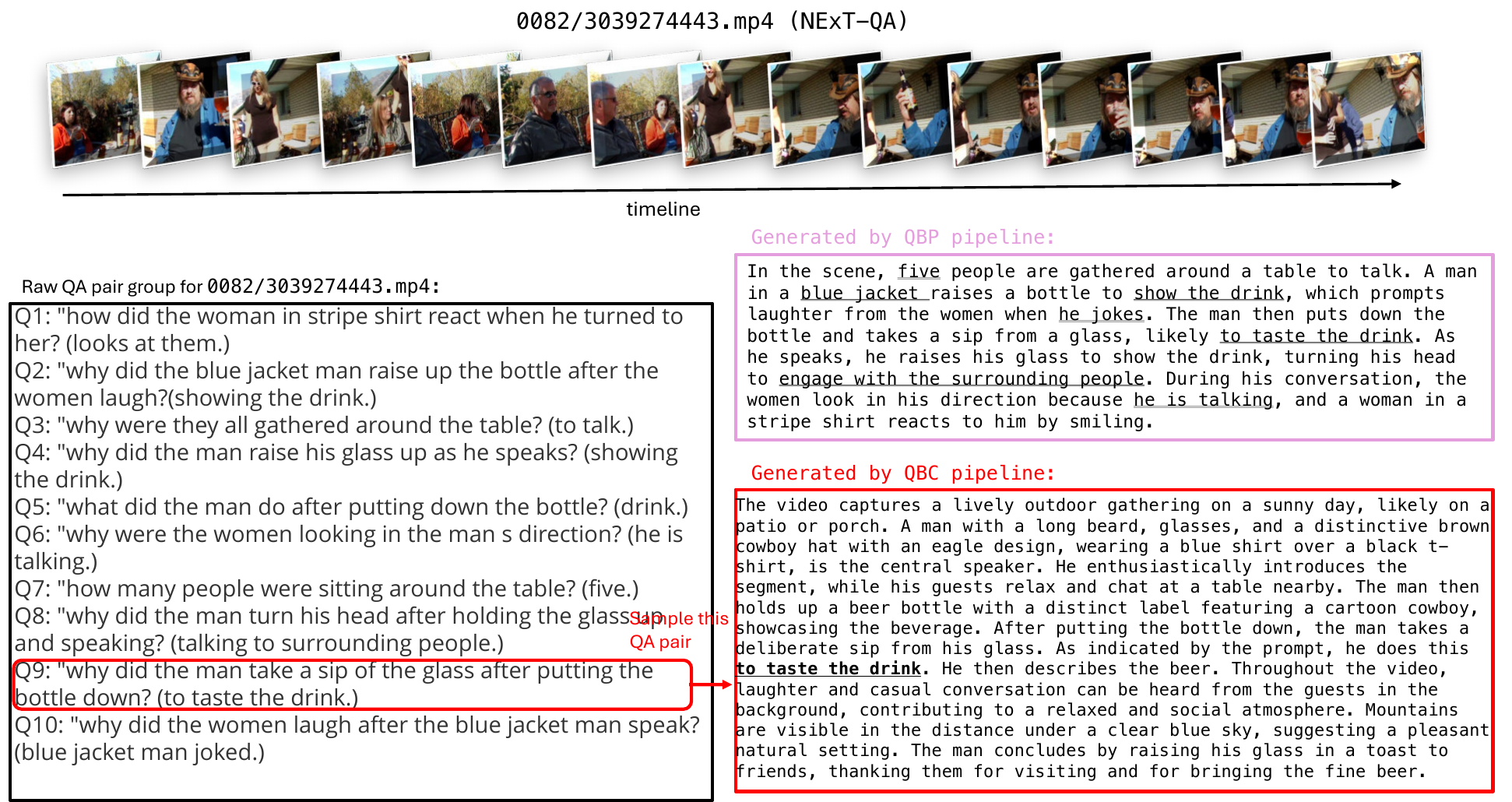}
    \caption{An example of synthesized data generated by our framework. The left shows the raw QA pairs from NExT-QA, while the right presents the corresponding outputs: a narrative produced by the QBP pipeline and rationales generated by the QBC pipeline.}
    \label{fig-app:qbp_qbc_example1}
\end{figure}

\begin{figure}[t]
    \centering
    \includegraphics[width=\linewidth]{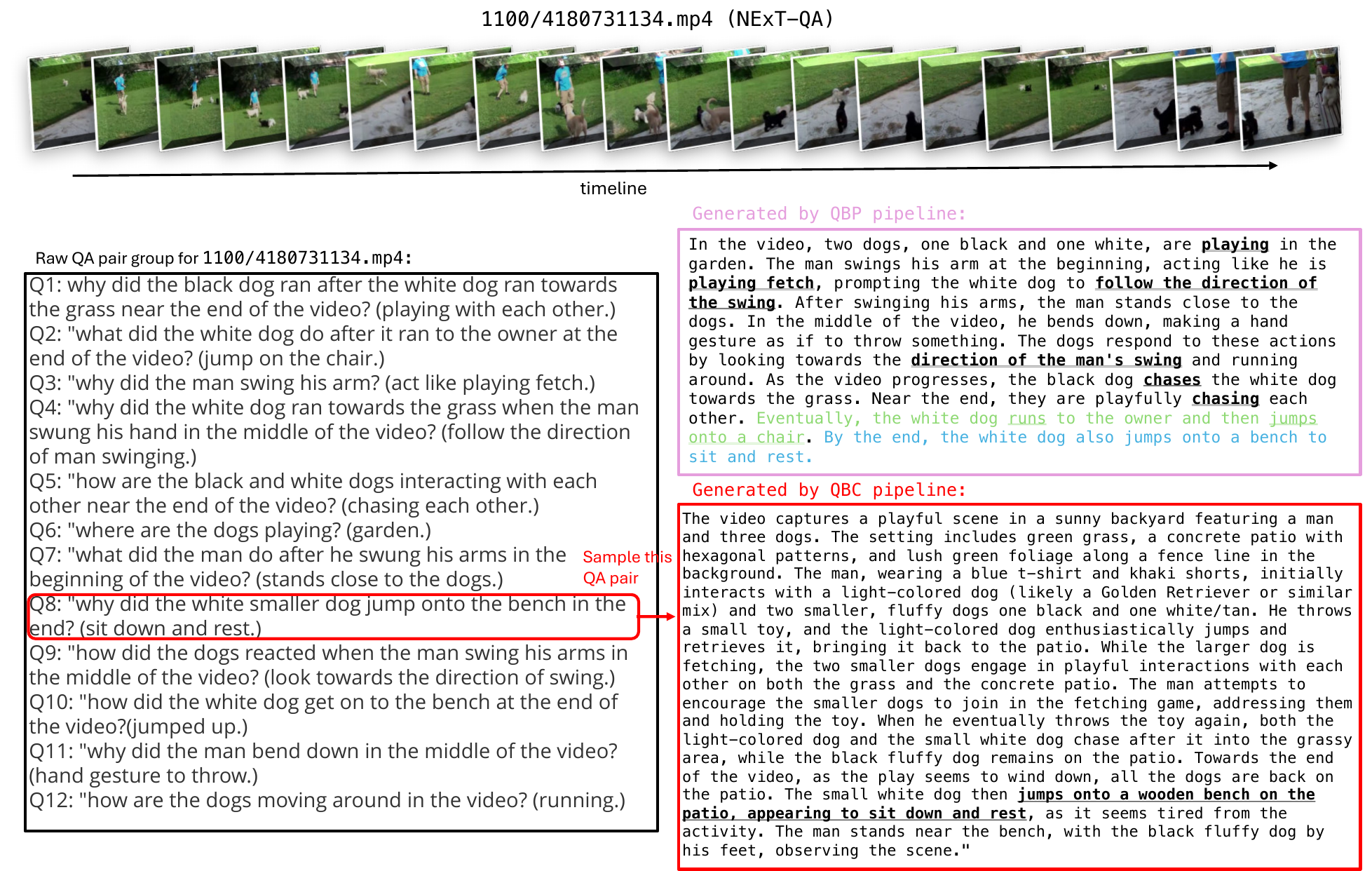}
    \caption{An example of synthesized data generated by our framework. The left shows the raw QA pairs from NExT-QA, while the right presents the corresponding outputs: a narrative produced by the QBP pipeline and rationales generated by the QBC pipeline. Due to overlapping content in the question group (e.g., Q2 and Q10), the generated QBP narrative includes two similar concluding sentences, marked in \textcolor{skyblue}{blue} and \textcolor{lightgreen}{green} for clarity, reflecting minor redundancy introduced by semantically repetitive QA pairs.}
    \label{fig-app:qbp_qbc_example2}
\end{figure}

To offer a more concrete understanding of our approach, this section showcases several examples generated by our proposed QBP and QBC methods. As illustrated in Figures~\ref{fig-app:qbp_qbc_example1} and~\ref{fig-app:qbp_qbc_example2}, these examples highlight the practical output and effectiveness of our techniques. 
As noted in Section~\ref{sec:error}, while the proportion of imperfections remains small, occasional issues are unavoidable. For instance, in Figure~\ref{fig-app:qbp_qbc_example2}, semantic overlap within the question group (e.g., Q2 and Q10) causes the QBP-generated narrative to include redundant concluding sentences.

% ====================================================================
% Appendix Section for Human Evaluation Protocol
% You can copy and paste this entire block into your appendix.
% ====================================================================
\section{Assessing the Quality of Synthetic Data}\label{app:assess}

\subsection{Human Evaluation}
\label{app:human_eval}

To quantitatively assess the quality of our synthesized data, we design and conduct a human evaluation study. We refer to QBP as Task A (Narrative Evaluation) and QBC as Task B (Rationale Evaluation). For each task, evaluators are asked to rate each generated text on a 1–5 scale across several quality dimensions, guided by the detailed descriptions provided. A score of 5 indicates the highest quality, while 1 indicates the lowest.

\begin{tcolorbox}[title=\textbf{1. Factual Consistency}, sharp corners]
    \textbf{Guiding Question:} Does the generated text contradict any of the facts provided in the source information (the QA pairs for Task A; the video and correct answer for Task B)?
    \begin{itemize}
        \item \textbf{5 (Excellent):} The text is perfectly consistent with all source facts.
        \item \textbf{3 (Moderate):} The text contains minor inaccuracies or makes claims that are plausible but not directly supported by the source.
        \item \textbf{1 (Poor):} The text directly contradicts a key fact from the source (e.g., says ``the person is running'' when the answer is ``walking'').
    \end{itemize}
\end{tcolorbox}

\begin{tcolorbox}[title=\textbf{2. Logical Coherence (Task A - QBP only)}, sharp corners]
    \textbf{Guiding Question:} Does the narrative describe events in a logical and coherent order? Does the story make sense?
    \begin{itemize}
        \item \textbf{5 (Excellent):} The sequence of events is clear, logical, and easy to follow. Causal and temporal relationships are sensible.
        \item \textbf{3 (Moderate):} The narrative is generally understandable, but the ordering of some events might be slightly awkward or ambiguous.
        \item \textbf{1 (Poor):} The narrative is confusing, jumbled, or illogical (e.g., describes an effect before its cause, or confuses the identities of different people).
    \end{itemize}
\end{tcolorbox}

\begin{tcolorbox}[title=\textbf{3. Visual Grounding (Task B - QBC only)}, sharp corners]
    \textbf{Guiding Question:} Does the rationale describe specific, observable evidence from the video that helps to justify the given answer?
    \begin{itemize}
        \item \textbf{5 (Excellent):} The rationale perfectly describes tangible visual details that serve as strong, direct evidence for the answer.
        \item \textbf{3 (Moderate):} The rationale is relevant but somewhat generic, describing the general scene rather than the specific evidence.
        \item \textbf{1 (Poor):} The rationale is irrelevant, describes something not visible in the video (fabrication), or simply rephrases the question without providing visual evidence.
    \end{itemize}
\end{tcolorbox}

\begin{tcolorbox}[title=\textbf{4. Fluency}, sharp corners]
    \textbf{Guiding Question:} Is the generated text well-written, grammatically correct, and easy for a native speaker to read?
    \begin{itemize}
        \item \textbf{5 (Excellent):} Flawless grammar and natural, fluent phrasing.
        \item \textbf{3 (Moderate):} Contains minor grammatical errors or awkward phrasing that do not impede understanding.
        \item \textbf{1 (Poor):} The text is ungrammatical, nonsensical, or very difficult to understand.
    \end{itemize}
\end{tcolorbox}

\end{document}